\documentclass[conference]{IEEEtran}

\IEEEoverridecommandlockouts
\usepackage{cite}
\usepackage{amsmath,amssymb,amsfonts}
\usepackage{algorithmic}
\usepackage{graphicx}
\usepackage{textcomp}
\usepackage{xcolor}
\usepackage[ruled,vlined]{algorithm2e}
\usepackage{stfloats}
\usepackage{enumitem}

\def\BibTeX{{\rm B\kern-.05em{\sc i\kern-.025em b}\kern-.08em
    T\kern-.1667em\lower.7ex\hbox{E}\kern-.125emX}}
\begin{document}

\title{Morphogenetic Assembly and Adaptive Control for Heterogeneous Modular Robots\\
}

\author{
    \IEEEauthorblockN{
        Chongxi~Meng$^{1}$,
        Da~Zhao$^{2}$,
        Yifei~Zhao$^{1}$,
        Minghao~Zeng$^{1}$,
        Yanmin~Zhou$^{1}$,
        Zhipeng~Wang$^{1,\dagger}$,
        Bin~He$^{1}$
    }
    \thanks{This work was supported in part by the National Natural Science Foundation of China under Grant 62088101, Grant 62403366 and Grant 62473294; in part by Shanghai Municipal Science and Technology Major Project under Grant 2021SHZDZX0100.}
    \thanks{$^{1}$Shanghai Research Institute for Intelligent Autonomous Systems, Tongji University, Shanghai, China.}
    \thanks{$^{2}$Department of Mechanical and Automation Engineering and T Stone Robotics Institute, The Chinese University of Hong Kong, Hong Kong.}
    \thanks{$^{\dagger}$Corresponding author Zhipeng~Wang: wangzhipeng@tongji.edu.cn}

}

\maketitle

\begin{abstract}
    This paper presents a closed-loop automation framework for heterogeneous modular robots, encompassing the entire pipeline from morphological construction to adaptive control. Within this framework, a mobile manipulator manipulates heterogeneous functional modules—including structural, joint, and wheeled modules—to dynamically assemble diverse robot configurations and grant them immediate locomotion capabilities. To address the state-space explosion inherent in large-scale heterogeneous reconfiguration, we propose a hierarchical planner: the high-level planner employs a bi-directional heuristic search with type penalty terms to generate module-handling sequences, while the low-level planner utilizes A* search to compute optimal execution trajectories. This approach effectively decouples discrete configuration planning from continuous motion execution. For adaptive motion generation of unknown assembled configurations, we introduce a GPU-accelerated Annealing Variance Model Predictive Path Integral (MPPI) controller. By incorporating a multi-stage variance annealing strategy to balance global exploration and local convergence, the controller achieves configuration-agnostic, real-time motion control. Large-scale simulations demonstrate that the type penalty term is crucial for planning robustness in heterogeneous scenarios. Furthermore, the greedy heuristic generates plans with lower physical execution costs compared to the Hungarian heuristic. The proposed Annealing-variance MPPI significantly outperforms standard MPPI in both velocity tracking accuracy and control frequency, achieving real-time control at 50 Hz. The framework successfully validates the full-cycle process, including module assembly, robot merging and splitting, and dynamic motion generation.

\end{abstract}


\section{Introduction}
Traditionally, the boundaries that separate robotics, architecture, and materials science have been gradually dissolving.
Automated construction has long been a goal in robotics, enabling the assembly and maintenance of uncrewed structures in extreme environments such as space exploration and post-disaster scenarios~\cite{petersen2019review}.
However, the ultimate ambition of building uncrewed structures goes far beyond erecting passive houses or bridges; it aims at the automated construction of large-scale functional systems that can sense, respond, and act---that is, robotic systems themselves.
Modern electric vehicles are essentially mobile robots composed of wheels and a body, while certain space stations with self-repair capabilities act as long-lived large robotic entities.
With the rapid progress of embodied intelligence, conventional architectural structures such as houses and bridges will also be viewed as potential robotic agents.
Therefore, the ultimate goal of this field is to realize the automated construction, reconfiguration, mobility, and intelligentization of these embodied functional structures, marking an evolution of embodied intelligence toward a new level.

\begin{figure}[t]
    \centering
    \includegraphics[width=\columnwidth]{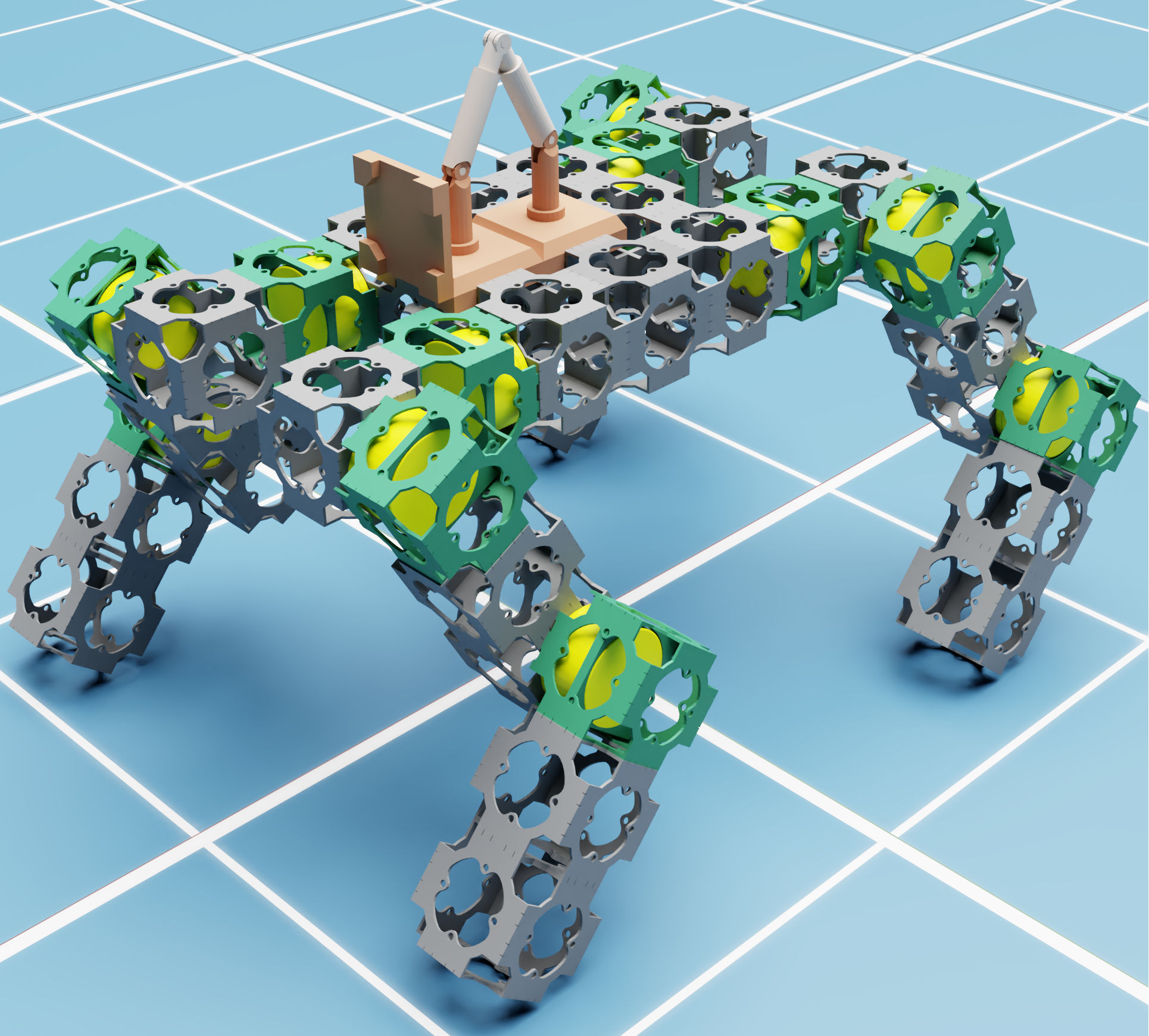}
    \caption{\textbf{Heterogeneous Modular Robotic System.} The system comprises a builder robot and a suite of standardized functional modules, including rigid units and actuated joint units. The figure demonstrates a quadrupedal configuration where the builder, integrated onto the torso, actively manipulates modules to reconfigure the body structure.}
    \label{fig:concept}
\end{figure}

This evolution challenges the widely adopted assumption of a \textbf{given body} in embodied intelligence.
Pursuing an intelligence that can self-create, self-evolve, and self-maintain may be the next frontier, because it directly points to a fundamental distinction between biological and artificial systems: the difference between \textbf{autopoietic systems} and \textbf{allopoietic systems}.
Machine life should exhibit morphological metabolism and adaptive self-maintenance of core functions~\cite{Wyder2024RobotMT, whitman2023learning}.
From the perspectives of physical assembly and adaptive control, we review current progress in the field and motivate the framework proposed in this work.

Existing progress in physical assembly can be broadly categorized into two paradigms: \textbf{self-reconfiguration} and \textbf{external assembly}.
Self-reconfiguration emphasizes embedding all intelligence and electromechanical components within each individual module; each self-reconfigurable modular robot is a complete agent on its own.
Whether by integrating self-soldering connectors, wheels, and magnetic couplers into homogeneous systems~\cite{neubert2016soldercubes, zhao2024snail, zhao2022snailbot}, or by developing complex heterogeneous systems with active struts and advanced sensing~\cite{Spinos2017VariableTT, tu2022freesn, tu2023configuration}, these approaches typically incur extremely high unit complexity, cost, and weight.
Thus, while effective for creating small multifunctional robots, this inherent complexity fundamentally conflicts with the organizational and functional characteristics of living systems.
External assembly instead employs simpler, often passive, modules that are manipulated by dedicated external robots.
This concept has been demonstrated by a series of systems that use mobile assemblers and advanced algorithms to construct large passive structures~\cite{terada2008automatic, abdel2022self, gregg2024ultralight}.
However, these systems mainly assemble structural skeletons for buildings, and the resulting structures themselves lack agency and intelligence.
Inspired by both mechanisms, as shown in Fig.~1, we propose a novel modular robotic system composed of carrier robots and functional modules, capable of dynamically assembling into multiple types of robotic embodiments.

\begin{figure}[t]
    \centering
    \includegraphics[width=\columnwidth]{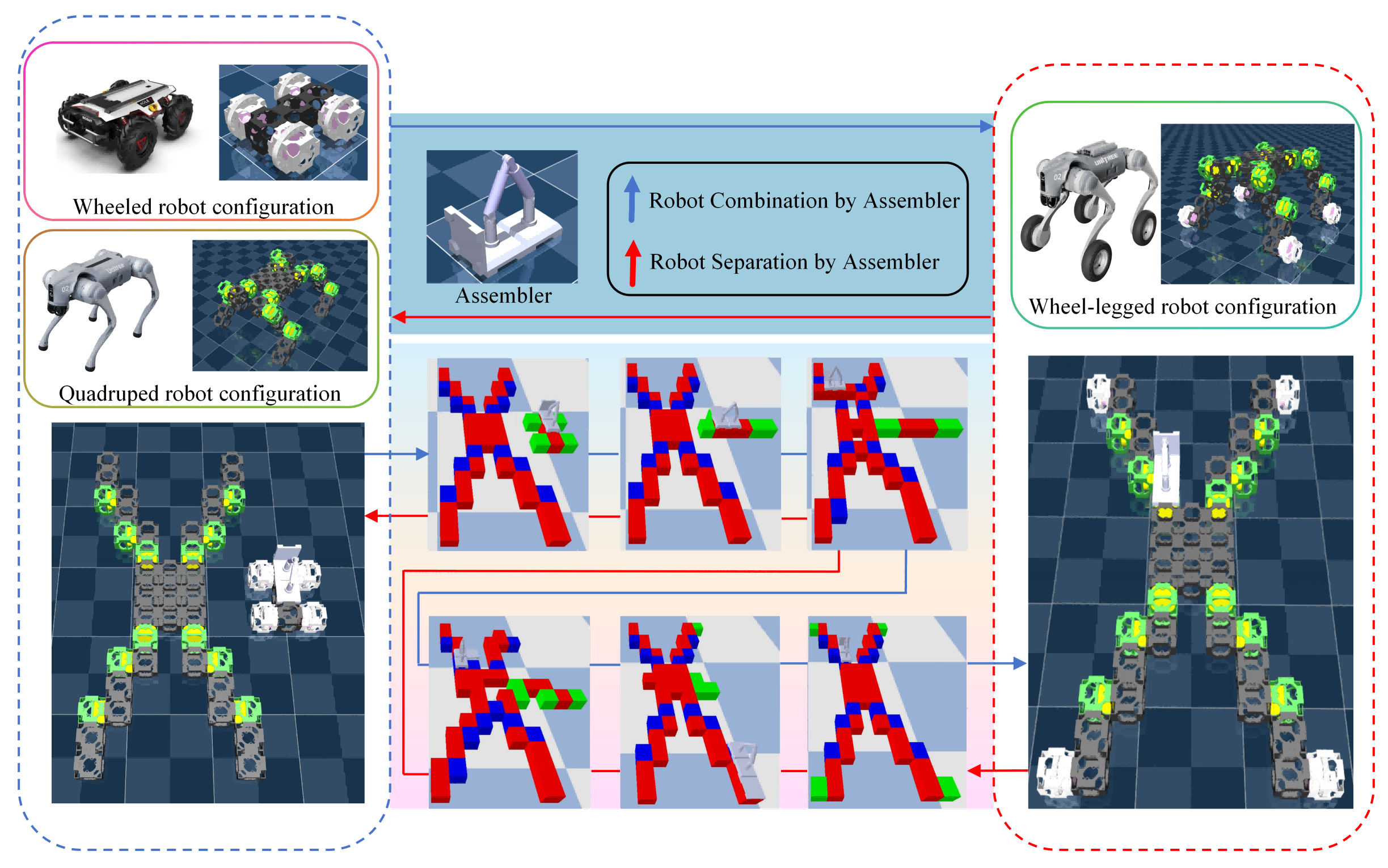}
    \caption{\textbf{Heterogeneous Modular Robotic System.}The framework enables dynamic morphological transitions between different robot configurations. \textbf{(Blue Arrows) Robot Combination:} An assembler robot merges a quadruped and a wheeled robot into a unified wheel-legged system by manipulating functional modules. \textbf{(Red Arrows) Robot Separation:} The inverse process where the composite wheel-legged robot is split back into its constituent quadruped and wheeled sub-systems. The center insets illustrate the step-by-step module manipulation sequence executed by the assembler.}
    \label{fig:system_pipeline}
\end{figure}

The enormous morphological space brings challenges for adaptive control, whose core difficulty lies in the large manifold gap.
Three lines of work provide answers from different angles: \textbf{cross-embodiment locomotion}, \textbf{morphology--control co-design}, and \textbf{evolutionary robotics}.
For cross-embodiment locomotion, the SMP strategy successfully controls multiple heterogeneous morphologies with a single policy by reusing control modules across different limbs and leveraging local message passing~\cite{huang2020smp, chen2023sgrl}.
A unified representation based on morphology--task graphs, together with a behavior distillation framework, effectively enables a single policy to generalize across multiple morphologies and tasks~\cite{furuta2023asystem, chen2024smerl}.
For morphology--control co-design, Robogrammar adopts MPPI as its underlying optimal controller and synthesizes stable gaits across diverse terrains and robot topologies, enabling performance evaluation of diverse morphologies under complex-terrain tasks~\cite{zhao2020robogrammar,Xu2021MultiObjectiveGH}.
End-to-end methods combining GNNs and Transformers encode robot topology and generate parameters of an action distribution through global information exchange over the body, from which concrete control actions are sampled~\cite{lu2025bodygen,yuan2022transform2act}.
In evolutionary robotics, the Triangle of Life theory combines offline evolution and online learning; using the RevDEknn algorithm, it adjusts CPG weights during an infant stage to adapt to the body morphology~\cite{eiben2020if,luo2022effects}.
State-of-the-art cross-embodiment robots often perform well on a single task such as locomotion, but when the action distributions induced by different embodiments differ too much, it becomes difficult for one distribution to fit multi-task capabilities~\cite{furuta2023asystem, chen2024smerl, bohlinger2024onepolicy}.
Therefore, following the explicit online MPPI idea in Robogrammar, we adopt MPPI as the controller for motion generation, and further optimize controller performance based on recent advances in GPU-accelerated simulated annealing~\cite{xue2025full}.

In summary, our framework addresses two identified core challenges:
\textbf{Morphogenetic construction:} we resolve the challenge of efficient heterogeneous assembly via a hierarchical planner. Acting as the execution engine of morphological reconfiguration, it translates high-level goals into a sequence of concrete physical construction steps.
\textbf{Adaptive Motion:} we address the need for adaptive control through a morphology-agnostic motion generation method. This controller endows any newly created morphology with the ability to act purposefully in the environment.

Our main contributions are as follows:
\begin{itemize}
    \item We propose a novel reconfigurable modular robotic framework that enables a single builder robot to autonomously assemble and actuate heterogeneous modular robots.
    \item We develop a hierarchical planner that efficiently solves complex heterogeneous assembly by decoupling high-level task planning from low-level motion execution.
    \item We introduce a morphology-agnostic, GPU-accelerated MPPI controller that provides immediate agency to novel morphologies.
    \item We validate our approach in simulation on complex tasks, including robot assembly, merging and splitting, as well as motion generation for the resulting morphologies.
\end{itemize}

\section{Problem Formulation}
\label{sec:problem_formulation}

\subsection{System Description}

\begin{figure}[t]
    \centering
    \includegraphics[width=\columnwidth]{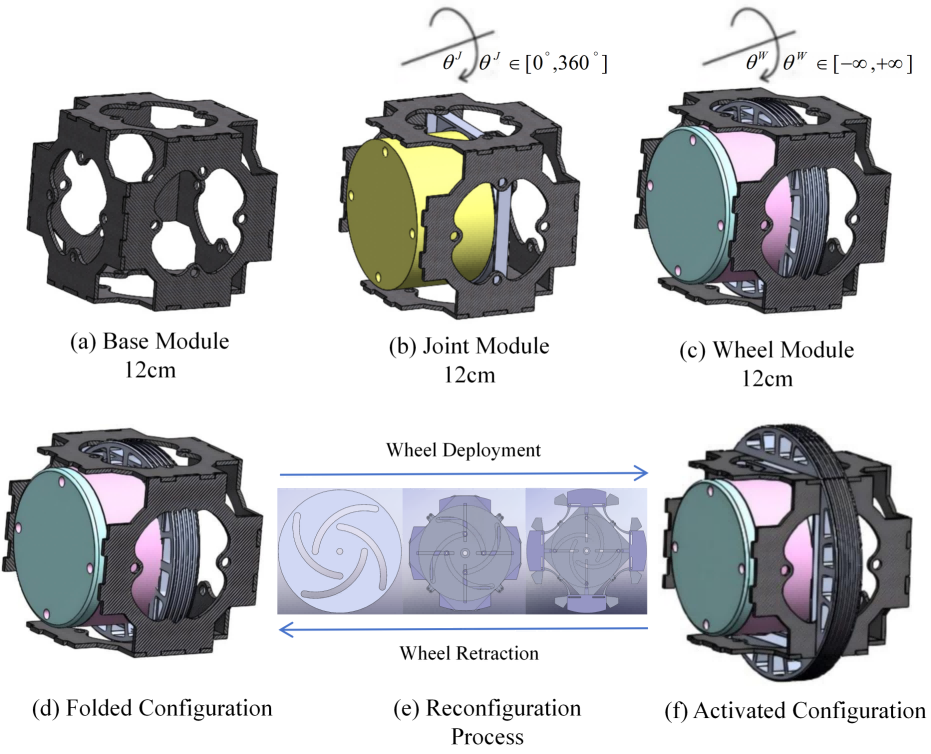}
    \caption{The three heterogeneous module types and the deployment mechanism of the wheel module:
        (a) The passive base module serves as the primary structural element.
        (b) The joint module provides a single rotational degree of freedom (DOF), $\theta^J \in [0^\circ, 360^\circ]$.
        (c) The wheel module enables continuous rotation ($\theta^W$) for locomotion.
        (d) The wheel in its retracted configuration.
        (e) The screw-driven reconfiguration process.
        and (f) The wheel in its fully deployed configuration, ready for locomotion.
        All modules share a common $12\,\text{cm}$ cubic form factor.}
    \label{fig:modules}
\end{figure}

The system comprises heterogeneous cubic modules and a single assembler robot. All modules share a $12\,\text{cm}$ cubic form factor and utilize reversible self-soldering connectors for mechanically robust, electrically conductive coupling \cite{neubert2014self,smith2024self}.

This framework enables the construction of diverse morphologies and dynamic reconfiguration between arbitrary configurations. This includes merging separate robots or executing reverse separation tasks, as illustrated in Fig.~\ref{fig:system_pipeline}. These processes are physical manifestations of morphological adaptation, wherein the system actively alters its own topology. We particularly analyze these two processes, which can be decomposed into instances of a fundamental reconfiguration problem.

The robot combination process is decomposed into two phases. First, in the approach phase, the smaller robot reconfigures itself into a connected auxiliary structure and locomotes by relocating its end modules to dock with the larger robot. Once connected, the merge phase begins, where the combined assembly is treated as a single entity and reconfigured from this intermediate connected state to the final target morphology.

Conversely, the robot separation process is the inverse. It starts by reconfiguring the initial robot into a state comprising one of the target sub-robot's final shapes and a connected auxiliary structure. This auxiliary structure then detaches, moves away, and subsequently reconfigures itself into the second target robot.

Therefore, the execution of these complex, reconfiguration-driven tasks fundamentally relies on solving the same core challenge: efficiently reconfiguring a connected set of modules from one configuration to another. Our primary focus is thus on developing a planner that excels at this core reconfiguration problem.

Three module types enable diverse robot morphologies (Fig.~\ref{fig:modules}): (a) Base modules provide structural support, (b) Joint modules incorporate fixed-axis motors enabling continuous rotation $\theta^J \in [0^\circ, 360^\circ]$, and (c) Wheel modules feature retractable wheels deployed via lead screw actuation and secured by electromagnetic locks. When retracted, wheel modules maintain cubic geometry for lattice compatibility; when deployed, they provide locomotion capability.

The modular structure configuration is defined as:
\begin{equation}
    \label{eq:config}
    C = (C^{1}, C^{2}, \dots, C^{M})
\end{equation}
where $M$ is the number of module types and $C^{m}$ specifies lattice coordinates occupied by modules of type $m$.

\begin{figure}[t]
    \centering
    \includegraphics[width=\columnwidth]{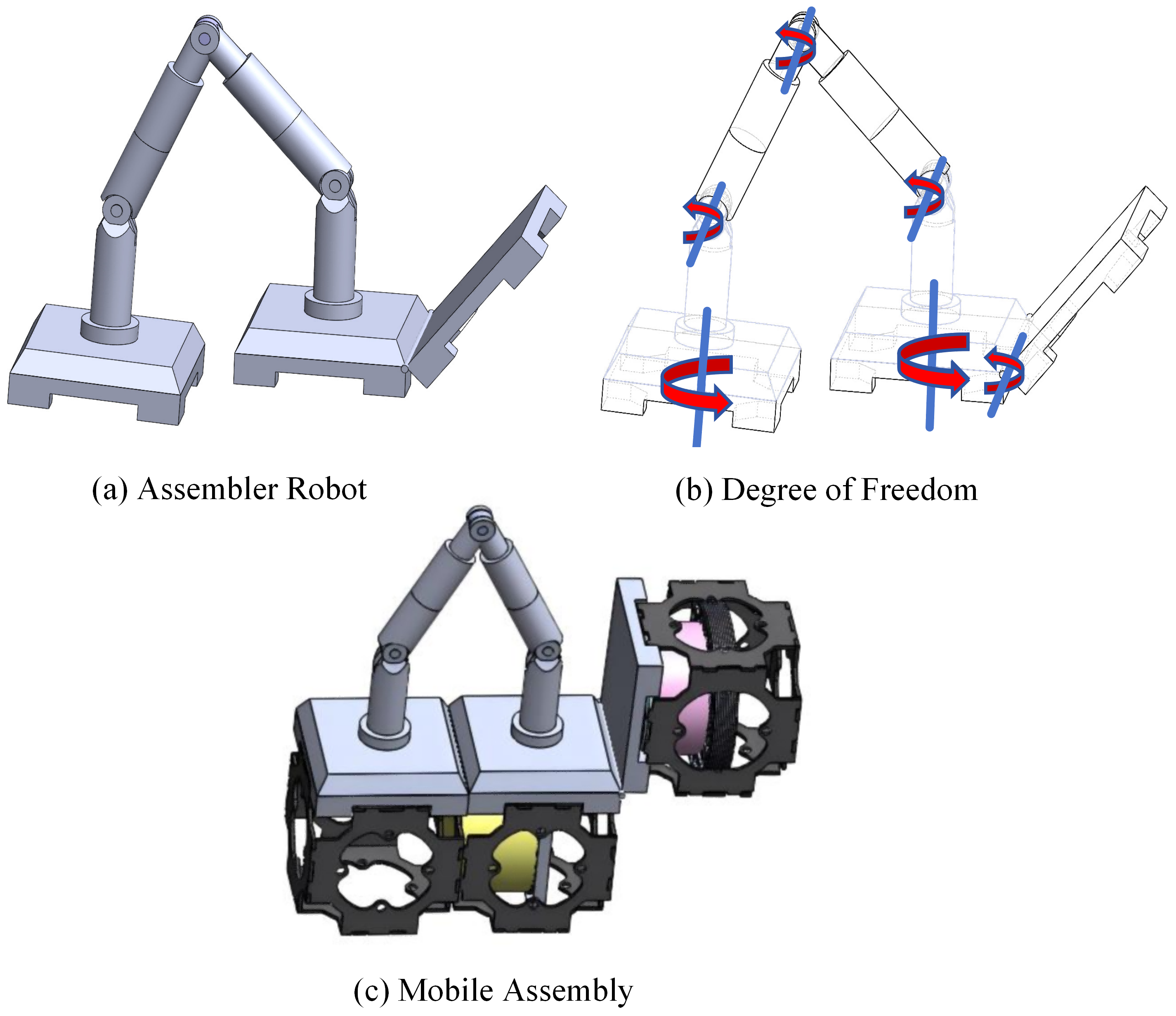}
    \caption{The assembler robot. (a) Overall design, featuring a dual-foot base and a multi-DOF manipulator. (b) The degrees of freedom (DOF) of the robot. (c) Demonstration of a mobile assembly task, where the robot locomotes across an assembled structure to manipulate a module.}
    \label{fig:assembler}
\end{figure}

The assembler robot (Fig.~\ref{fig:assembler}) features a dual-foot base for surface traversal and a multi-DOF manipulator with a claw-like end-effector. The base's four protrusions align with module surface recesses, enabling reliable attachment. By alternately engaging its feet, the robot locomotes across assembled structures while manipulating individual modules.

The robot state is described by:
\begin{equation}
    \label{eq:robot_state}
    R = (l, m_{\mathrm{held}})
\end{equation}
where $l$ represents end-effector location and $m_{\mathrm{held}}$ denotes the held module.

The action space $\mathcal{A}_{\text{robot}}$ includes:
\begin{itemize}
    \item \textit{Locomotion}: Moving one foot to adjacent, unoccupied cells while the other remains fixed.

    \item \textit{Rotation}: Pivoting $\pm90^\circ$ or $180^\circ$ around either foot to reposition on the structure surface.

    \item \textit{Manipulation}: Pick and place operations that preserve structural connectivity when retrieving or depositing modules.
\end{itemize}

\subsection{Problem Statement}

The first problem addresses the executive mechanics of autonomous assembly by formalizing the process of morphogenesis as a planning problem:

\textbf{Highly Efficient Reconfiguration Problem:}
The objective is to find an action sequence
\begin{equation}
    A = (a_1,\dots,a_T), \quad a_t \in \mathcal{A}_{\mathrm{robot}}
\end{equation}
such that the final state satisfies
\begin{equation}
    G(C_{\mathrm{final}}) \cong G(C_{\mathrm{goal}}), \quad R_{\mathrm{final}} = R_{\mathrm{goal}}
\end{equation}
where $G(\cdot)$ denotes the connectivity graph of the structure, and $C_{\mathrm{final}}, R_{\mathrm{final}}$ are the states after executing $A$.

The cost of a plan is defined as
\begin{equation}
    \mathrm{Cost}(A) = \sum_{t=1}^T \mathrm{cost}(a_t)
\end{equation}
where each action cost corresponds to the minimal execution cost derived from low-level motion planning. The efficient plan is then given by
\begin{equation}
    \label{eq:optimality}
    A^* = \arg\min_A \mathrm{Cost}(A)
\end{equation}
subject to
\begin{equation}
    G(C_{\mathrm{final}}) \cong G(C_{\mathrm{goal}})  \quad R_{\mathrm{final}} = R_{\mathrm{goal}}
\end{equation}

The second problem addresses the control challenge inherent to autonomous adaptation: enabling a system to effectively operate its newly formed morphology.

\textbf{Adaptive Motion Problem:}
Given a high-level task command, such as a target velocity $\mathbf{v}_{\mathrm{cmd}}$, the objective is to find a control policy $\pi$ that generates actions $a_t$ from states $s_t$ to maximize the expected cumulative reward over a finite horizon $H$:
\begin{equation}
    \label{eq:motion_objective}
    \pi^* = \arg\max_{\pi} \mathbb{E} \left[ \sum_{t=0}^{H-1} R(s_t, a_t; \mathbf{v}_{\mathrm{cmd}}) \right]
\end{equation}
subject to the robot's unknown dynamics $s_{t+1} = f(s_t, a_t)$. A critical constraint of this problem is that the method of finding $\pi^*$ must be morphology-agnostic, which must apply to any valid assembled robot defined by $G_{\mathrm{goal}}$ without requiring architecture-specific retraining or manual design.

\section{Hierarchical Reconfiguration Planner}
\label{sec:methods}
The high-dimensional nature of the reconfiguration problem, spanning both discrete module placements and continuous robot motions, necessitates an efficient planning strategy. We address this with a hierarchical framework that decouples high-level task planning from low-level motion execution, as outlined in Algorithm~\ref{alg:hierarchical}. This decomposition allows for a tractable search in the vast configuration space while retaining a fine-grained evaluation of the final plan's execution cost.

The high-level planner operates in the discrete configuration space of the modular structure. It employs a bidirectional heuristic search to find a sequence of module relocations that transforms the initial configuration $C_{\mathrm{ini}}$ into the goal $C_{\mathrm{goal}}$. The search is guided by a heuristic function $h(S_C)$ that estimates the number of remaining relocations. We investigate two primary heuristics: one derived from the Hungarian algorithm, which provides an admissible estimate by solving the optimal assignment problem between misplaced modules and their targets, and a faster, non-admissible Greedy alternative. To effectively handle heterogeneous assemblies, these heuristics are augmented with a type-penalization term to break search symmetries and provide stronger directional guidance. The validity of each macro-action is enforced by ensuring any picked module is a leaf node whose removal does not partition the structure's connectivity graph via graph traversal, and that the drop-off location is within the assembler's kinematic reach.

For each module relocation macro-action proposed by the high-level planner, the low-level planner computes a cost-optimal execution trajectory for the assembler robot. This is formulated as an A* search in the robot's state space, which is defined by its foot placements and any held module. The objective is to find a sequence of primitive actions, including locomotion, rotation, and manipulation, that minimizes the total execution cost. This cost, $\mathrm{Cost}(\tau)$, is the sum of the costs of these primitive actions and directly corresponds to the physical effort and time required for the task, consistent with the optimality criterion in Eq.~\ref{eq:optimality}.

\begin{algorithm}[t]
    \small
    \DontPrintSemicolon
    \caption{Bidirect-Hierarchical Reconfiguration with Heuristic $h$}
    \label{alg:hierarchical}
    \KwIn{$C_{\mathrm{ini}},\,C_{\mathrm{goal}}$}
    \KwOut{Executable plan $\mathcal{A}=\{(A_H,\tau)\}$}
    Initialize priority queues $Q_{\mathrm{fwd}},Q_{\mathrm{bwd}}$ keyed by $f=g+h$\;
    Push $(C_{\mathrm{ini}},g{=}0)$ into $Q_{\mathrm{fwd}}$; push $(C_{\mathrm{goal}},g{=}0)$ into $Q_{\mathrm{bwd}}$\;
    $\text{Parent}_{\mathrm{fwd}},\text{Parent}_{\mathrm{bwd}}\leftarrow\emptyset$; $\text{Visited}_{\mathrm{fwd}},\text{Visited}_{\mathrm{bwd}}\leftarrow\emptyset$\;
    \While{$Q_{\mathrm{fwd}}\neq\emptyset$ and $Q_{\mathrm{bwd}}\neq\emptyset$}{
    $\mathcal{D}\leftarrow \arg\min_{\mathrm{dir}\in\{\mathrm{fwd},\mathrm{bwd}\}} |Q_{\mathrm{dir}}|$
    $(S,g)\leftarrow Q_{\mathcal{D}}.\mathrm{pop}()$; add $S$ to $\text{Visited}_{\mathcal{D}}$\;
    \If{$S \in \text{Visited}_{\mathrm{fwd}}\cap\text{Visited}_{\mathrm{bwd}}$}{\textbf{break} and join paths to form $\Pi_H$}
    \ForEach{$A_H \in \mathrm{GenerateRelocations}(S)$}{
        \If{$\mathrm{IsLeafPickup}(A_H)$ and $\mathrm{MaintainsConnectivity}(S,A_H)$ and $\mathrm{ReachableDrop}(A_H)$}{
            $S'\leftarrow \mathrm{Apply}(S,A_H)$; $g'\leftarrow g+1$; $h'\leftarrow h(S')$\;
            \If{$S'$ not seen in $\mathcal{D}$ or $g'$ improves}{
                record parent; push $(S',g')$ with key $g'+h'$ into $Q_{\mathcal{D}}$\;
            }
        }
    }
    }
    $\Pi_H\leftarrow \mathrm{ReconstructMacroPlan}(\text{Parent}_{\mathrm{fwd}},\text{Parent}_{\mathrm{bwd}})$; $\mathcal{A}\leftarrow \emptyset$\;
    \ForEach{$A_H\in\Pi_H$}{
    $\mathcal{P}\leftarrow \mathrm{TaskPoses}(A_H)$
    $(J^\star,\tau^\star)\leftarrow(\infty,\emptyset)$\;
    \ForEach{$p\in\mathcal{P}$}{
    $\tau\leftarrow \mathrm{AStarRobot}\big(S_R^{\mathrm{start}}(A_H),\,p,\,h=\mathrm{Manhattan}\big)$\;
    \If{$\mathrm{Cost}(\tau)<J^\star$}{$(J^\star,\tau^\star)\leftarrow(\mathrm{Cost}(\tau),\tau)$}
    }
    append $(A_H,\tau^\star)$ to $\mathcal{A}$\;
    }
    \Return $\mathcal{A}$\;
\end{algorithm}

\section{Dynamic Motion Generation}
\label{sec:motion_generation}
Upon completing assembly, the framework must endow the resulting morphology with dynamic locomotion. This is achieved through a two-stage pipeline: first, the robot's final configuration graph $G_{\mathrm{goal}}$ is automatically translated into a simulatable model $\mathcal{M}$ for a GPU-accelerated physics engine. Second, a morphology-agnostic motion controller generates control actions online.

Motion generation is framed as a finite-horizon optimal control problem to find an action sequence $A^* = \{a_0,\dots,a_{H-1}\}$ that maximizes a cumulative reward:
\begin{equation}
    A^* = \arg\max_{A} \mathbb{E} \left[ \sum_{t=0}^{H-1} R(s_t, a_t) \right],
\end{equation}
where the dynamics $s_{t+1} = f(s_t, a_t)$ are implicitly provided by the simulator. We employ a sampling-based Model Predictive Control strategy, which is well-suited for this task as it requires no analytical dynamics model and is highly parallelizable.

To achieve the real-time performance necessary for dynamic locomotion, we leverage massively parallel GPU-based simulation. This contrasts with prior work in modular robotics, such as RoboGrammar \cite{zhao2020robogrammar}, which relied on CPU-based MPPI and was thus limited to offline policy optimization. However, standard MPPI suffers from a fixed sampling variance, leading to a trade-off between exploration and exploitation that compromises performance on the complex, non-convex optimization landscapes of post-assembly morphologies.

To address this, we implement an enhanced MPPI controller inspired by recent work connecting sampling-based control with Annealed-Variance MPPI \cite{xue2025full}. Our controller employs a multi-stage \textit{annealing} strategy within each control cycle. It iteratively refines the action sequence by starting with a large sampling variance for global exploration and then gradually reducing it to achieve precise convergence. This approach, detailed in Algorithm~\ref{alg:annealed_mppi}, dynamically balances coverage and convergence, yielding robust and high-performance locomotion.

\begin{algorithm}[t]
    \caption{GPU-Accelerated Annealed MPPI}
    \label{alg:annealed_mppi}
    \small
    \DontPrintSemicolon
    Initialize nominal action sequence $U$\;
    \While{task not completed}{
    \For{iteration $i=1$ \KwTo $N$ (annealing steps)}{
    Compute annealed sampling covariance $\Sigma_i$ (decreases as $i$ increases)\;
    Sample $K$ noise sequences $\{\delta U_k\}_{k=1}^K \sim \mathcal{N}(0, \Sigma_i)$ in parallel on GPU\;
    Evaluate costs $S(U + \delta U_k)$ for all $k$ via parallel rollouts\;
    Compute weights $w_k \propto \exp(-S(U + \delta U_k)/\lambda)$\;
    Update nominal sequence: $U \leftarrow \sum_{k=1}^K w_k (U + \delta U_k)$\;
    }
    Execute first action $a_0$ from the refined sequence $U$\;
    Update $U$ for the next time step (receding horizon)\;
    }
\end{algorithm}

\section{Experiments}

To systematically evaluate our framework, all experiments were conducted on a workstation equipped with an i9-13900KF CPU and an NVIDIA 4090D GPU. We benchmarked our planner on a large suite of procedurally generated reconfiguration tasks, designed to span various dimensions of difficulty:
\begin{itemize}[leftmargin=*, nosep]
    \item \textbf{Problem Scale ($N$):} The total number of modules, ranging from 20 to 75, to test scalability.
    \item \textbf{Configuration Overlap:} Defined by the initial-goal configuration overlap, categorized as low (0.8), medium (0.5), and high (0.2).
    \item \textbf{Heterogeneous Composition:} Three distinct module type ratios were used to represent increasing degrees of heterogeneity:
          \begin{itemize}[leftmargin=1.5em, nosep]  % 手动调缩进
              \item \textbf{typeA} (Low): [100\% Base, 0\% Joint, 0\% Wheel]
              \item \textbf{typeB} (Medium): [80\% Base, 20\% Joint, 0\% Wheel]
              \item \textbf{typeC} (High): [60\% Base, 20\% Joint, 20\% Wheel]
          \end{itemize}
\end{itemize}
For each parameter combination, performance was measured over 100 random instances with a 60-second timeout per trial to ensure statistical significance.

\subsection{Ablation Study on Heuristic Design}

\begin{figure}[t]
    \centering
    \includegraphics[width=0.9\linewidth]{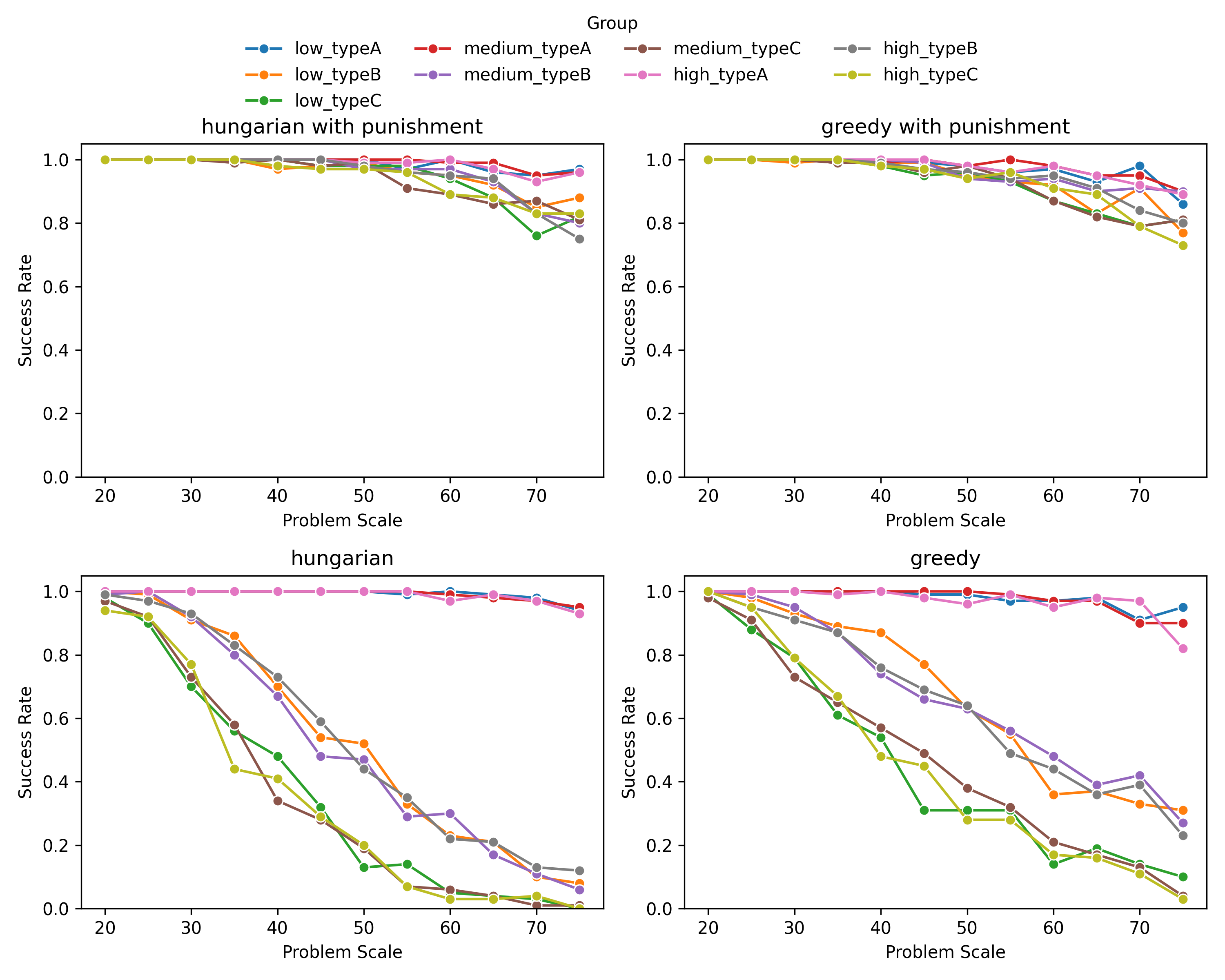}
    \caption{Success rate comparison for heuristics with and without the type-penalization term. The penalty is shown to be critical for ensuring planner robustness and scalability across our diverse benchmark suite.}
    \label{fig:penalty}
\end{figure}

To validate the necessity of the type-penalization term proposed in Section~\ref{sec:methods}, we conducted an ablation study on the benchmark suite. As shown in Fig.~\ref{fig:penalty}, heuristics with the penalty term maintained high success rates ($>80\%$) across nearly all scales and complexities. In stark contrast, performance without the penalty collapsed as problem scale and heterogeneity increased, with success rates approaching zero in the most challenging cases. This confirms that type-penalization is essential for breaking heuristic ties and avoiding symmetric deadlocks, thereby ensuring the planner's robustness and scalability for complex, multi-type tasks.

\subsection{Impact of Heuristics on Execution Cost}

Beyond planning success rates, we evaluated how the choice of high-level heuristic affects the final plan quality. We define plan quality by its \textit{total execution cost}: the cumulative cost of all primitive assembler actions (e.g., locomotion, rotation) computed by the low-level A* planner. This metric directly corresponds to the physical time and effort required for reconfiguration.

\begin{figure}[t]
    \centering
    \includegraphics[width=\columnwidth]{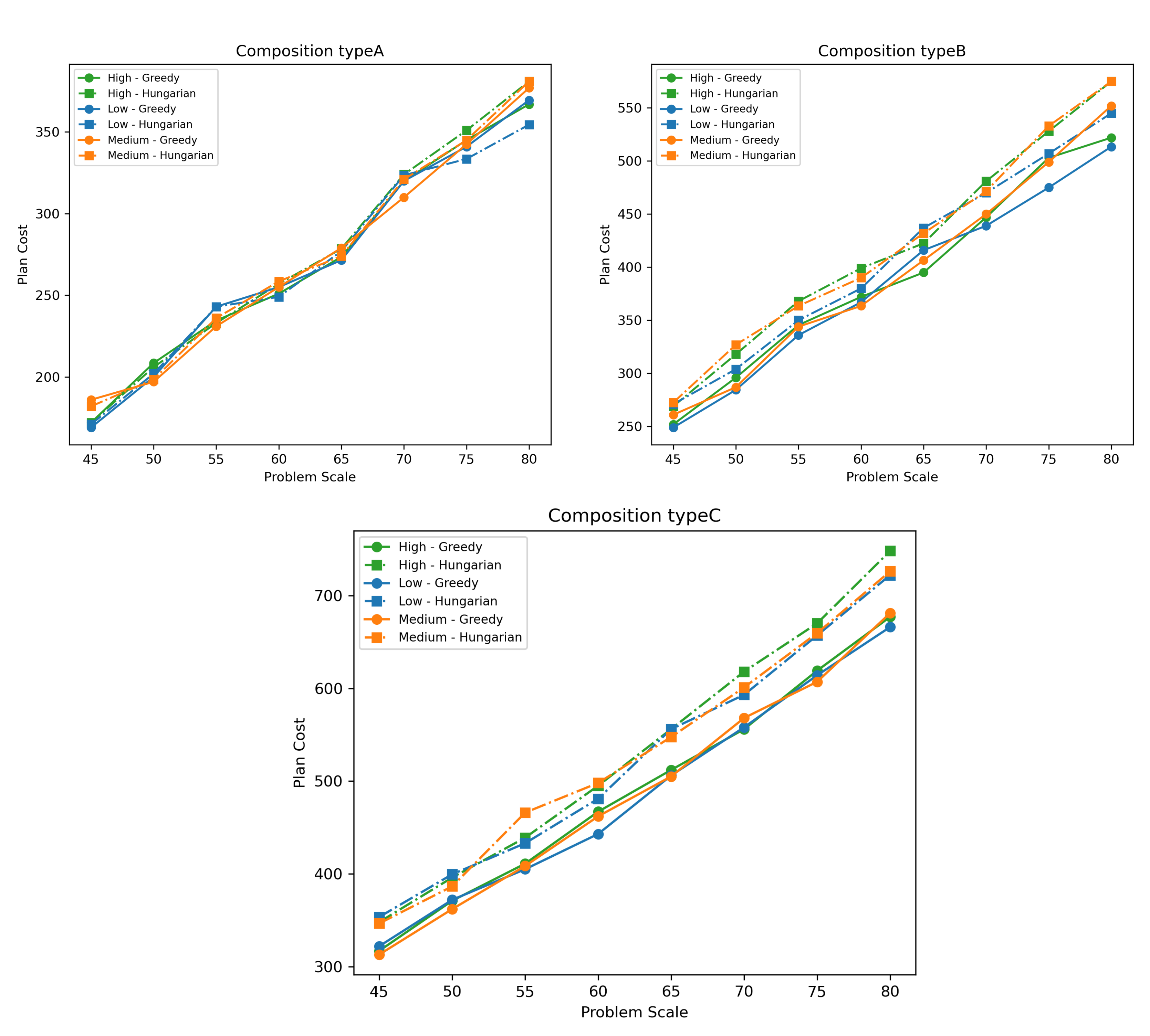}
    \caption{Comparison of the total plan execution cost for solutions generated using Greedy versus Hungarian heuristics. The Plan Cost represents the cumulative cost of all primitive robot actions. The Greedy heuristic consistently finds more economical plans, especially in highly heterogeneous scenarios (typeC).}
    \label{fig:plan_cost_comparison}
\end{figure}

As shown in Fig.~\ref{fig:plan_cost_comparison}, the Greedy heuristic consistently guides the planner to solutions with a lower total execution cost than the Hungarian heuristic. This cost-saving advantage becomes more pronounced as module heterogeneity increases (from typeA to typeC). While the Hungarian algorithm provides a globally optimal assignment in an abstract sense, the Greedy approach appears to generate a sequence of relocations that is more locally efficient for the assembler to execute, culminating in a more economical and practical physical plan.

\subsection{Controller Performance Evaluation}

\begin{figure}[!htbp]
    \centering
    \includegraphics[width=\columnwidth]{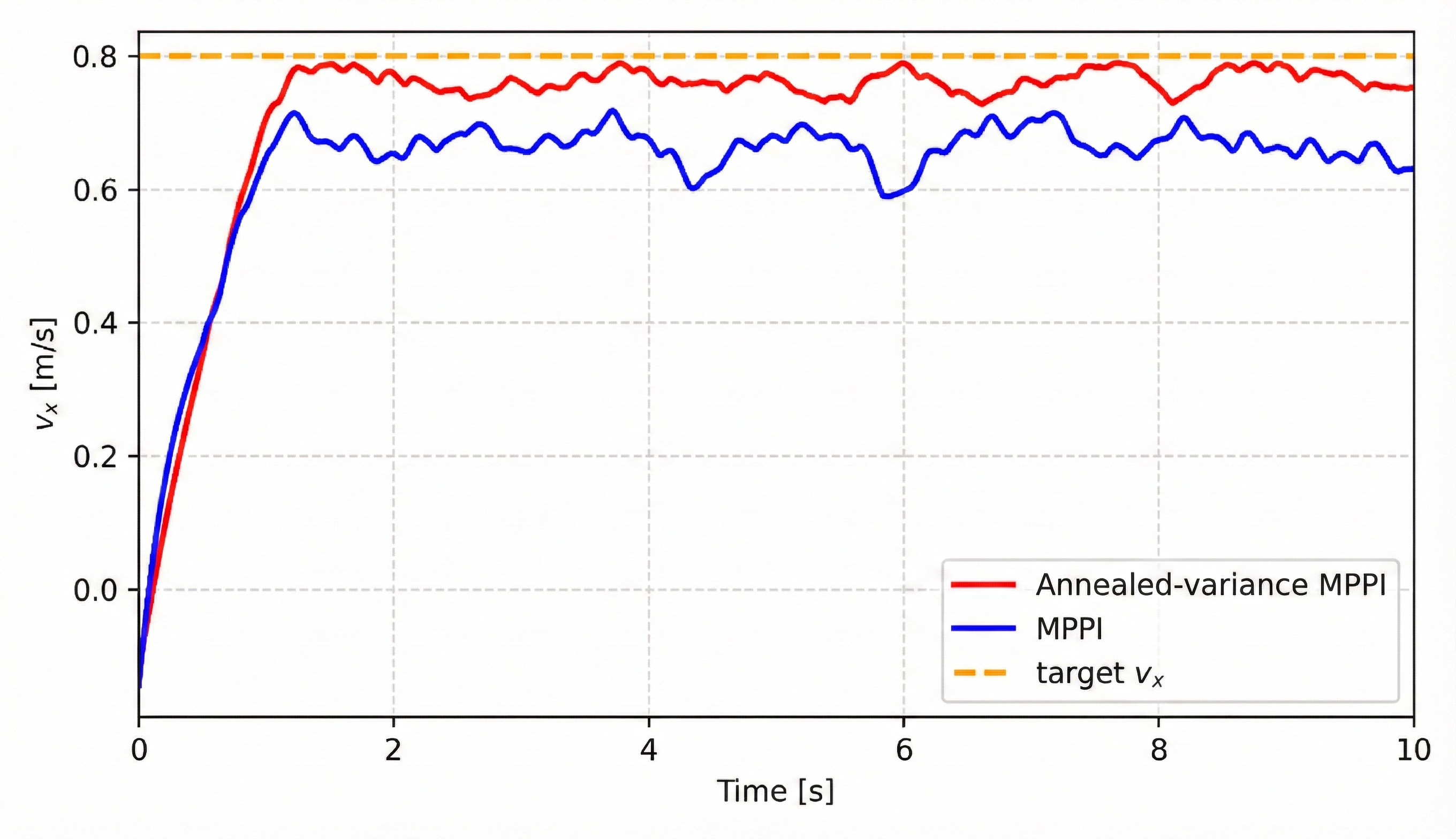}
    \caption{Comparison of velocity tracking performance between the standard MPPI and annealed-variance MPPI. The task is to track a target velocity of $v_x = 0.8\,\text{m/s}$. The annealed-variance MPPI (red) achieves significantly better tracking accuracy, closely matching the target, while the standard MPPI (blue) suffers from a persistent steady-state error and larger oscillations.}
    \label{fig:velocity_tracking_comparison}
\end{figure}

Finally, we evaluated the performance of the dynamic motion generation pipeline. We benchmarked our sampling-based Model Predictive Path Integral controller on a velocity tracking task using an assembled quadruped morphology. The objective was to track a target forward velocity of $v_x = 0.8\,\text{m/s}$. We compared the performance of a standard MPPI implementation against our proposed variant, which is inspired by annealed variance to improve sampling efficiency.

As depicted in Fig.~\ref{fig:velocity_tracking_comparison}, the results confirm our hypothesis. The annealed-variance MPPI (red) demonstrates substantially better tracking accuracy, rapidly converging to the target velocity with minimal steady-state error. In contrast, the standard MPPI (blue) exhibits a persistent velocity offset and significant oscillations. Furthermore, GPU-accelerated implementation of the annealed-variance MPPI achieves a 50\,Hz control frequency, meeting real-time requirements, whereas the standard MPPI was limited to approximately 3\,Hz on the same task. These results validate that integrating an advanced controller like annealed-variance MPPI into our framework yields high-performance, stable locomotion, effectively closing the loop from high-level assembly planning to adaptive dynamic control.

\begin{figure*}[t]
    \centering
    \includegraphics[width=\textwidth]{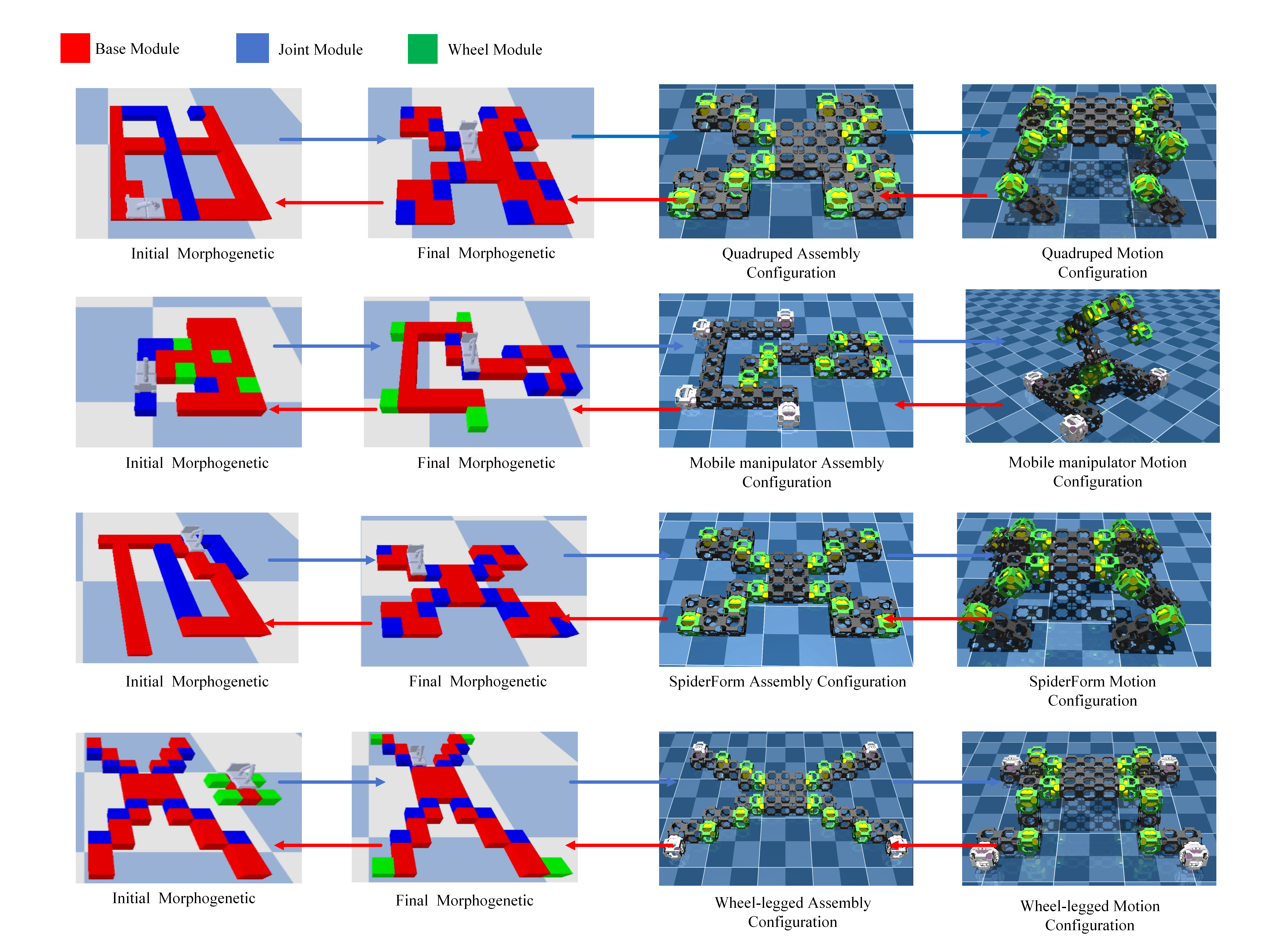} % 请替换为您的图片文件名
    \caption{
        Overall illustration of the integrated framework. The framework is formulated as a closed-loop workflow between (i) a modular, standardized resource workspace that stores and supplies hardware modules and (ii) a modular robot assembled from these modules. Each row presents one representative configuration produced by this workflow: (top to bottom) a quadruped, a mobile manipulator, a spider-like multi-legged robot, and a wheel--leg hybrid robot. The pipeline covers planning, module selection and composition, physical assembly, and control deployment.
        \newline
        \textbf{Blue Arrows (Workspace $\rightarrow$ Robot):} Forward reconfiguration (assembly). The system selects modules from the shared inventory and composes them into a task-oriented robot. The four examples demonstrate that the same module set and interfaces can support distinct designs: a locomotion-focused quadruped, a manipulation-capable mobile base, an alternative stable multi-legged layout, and a heterogeneous wheel--leg platform with expanded mobility.
        \newline
        \textbf{Red Arrows (Robot $\rightarrow$ Workspace):} Reverse reconfiguration (disassembly and return-to-inventory). The robot is decomposed into standardized modules which are returned to the workspace for later reuse. This enables rapid adaptation under changing task requirements; e.g., modules from the quadruped can be reassigned to form the mobile manipulator, while the wheel--leg platform can be disassembled to supply components for other configurations or multiple simpler robots.
        \newline
    }
    \label{fig:holistic_demo}
\end{figure*}

\section{Conclusion and Future Work}
This paper presents the development and implementation of a novel closed-loop automation framework for heterogeneous modular robots, spanning from reconfiguration planning to adaptive control. The core contributions of this framework are twofold: first, a hierarchical planner is introduced to mitigate the state-space explosion inherent in large-scale module assembly tasks, enabling highly scalable path planning; second, a GPU-accelerated Model Predictive Path Integral controller with an annealing mechanism is utilized to achieve morphology-agnostic, real-time motion generation. This ensures stable locomotion and manipulation across diverse configurations and tasks. Extensive simulations validate that the proposed pipeline successfully executes the entire workflow from high-level commands to low-level physical execution. Future research will focus on addressing Sim-to-Real transfer challenges to verify the robustness of these algorithms on physical hardware.

From an engineering perspective, this framework provides a comprehensive lifecycle management solution for modular robotic systems. By treating the collection of modules as a recyclable hardware resource pool, the system achieves optimal resource allocation through algorithmic scheduling.

During the deployment phase, the system automatically assembles the required physical configuration based on task specifications via a morphogenesis algorithm. This process essentially transforms a static assembly of modules into a functional unit with active mobility. In the operational phase, the system relies on adaptive control algorithms to maintain structural stability and execute specific tasks, while retaining the capability for real-time reconfiguration in response to environmental feedback or mission changes.

Upon task completion or during the resource recovery phase, a deconstruction procedure is executed. The specific robotic morphology is disassembled and its control logic is terminated, allowing the physical modules to be returned to the resource pool. This modular reuse mechanism significantly enhances system flexibility and hardware utilization efficiency. For instance, the system can disassemble a quadruped robot after a transport mission and repurpose the same modules to construct a mobile manipulator for subsequent assembly tasks, as illustrated in Fig.~\ref{fig:holistic_demo}.

\bibliographystyle{IEEEtran}
\bibliography{my_references}

\end{document}